\documentclass{bmvc2k}

\usepackage{times}
\usepackage{epsfig}
\usepackage{graphicx}
\graphicspath{ {./figures/} }
\usepackage{amsmath}
\usepackage{amssymb}
\usepackage{makecell}
\usepackage{color}
\usepackage{xcolor}
\usepackage{wrapfig}

\usepackage{booktabs}
\usepackage{multirow}
\usepackage{hyperref}

\usepackage{cleveref}

\title{SynThermFace: Amplifying Limited Paired Data for Visible–Thermal Face Recognition via Synthetic Data Generation}

\addauthor{Anjith George}{anjith.george@idiap.ch}{1}
\addauthor{Adam Unal}{adam.unal@idiap.ch}{1}
\addauthor{Sebastien Marcel}{sebastien.marcel@idiap.ch}{1,2}

\addinstitution{
Idiap Research Institute\\
Rue Marconi 19\\
Martigny, Switzerland
}

\addinstitution{
University of Lausanne (UNIL)\\
Lausanne, Switzerland
}

\runninghead{George, Unal, Marcel}{SynThermFace}

\begin{document}

\maketitle

\begin{abstract}
Face recognition (FR) is a widely used modality for biometric authentication, but conventional models rely on visible-spectrum imagery and degrade when high-quality RGB images cannot be captured. Cross-spectral face recognition addresses this limitation by matching visible images with other modalities such as thermal imagery, enabling more reliable performance in low-light, nighttime, and unconstrained conditions. However, progress is limited by the scarcity of paired visible–thermal data, which is difficult and costly to collect at scale. We propose SynThermFace, a framework that amplifies limited real visible–thermal supervision into larger paired adaptation datasets for cross-spectral face recognition. A diffusion model is first adapted using a limited set of paired visible--thermal images and then used to generate large-scale paired visible--synthetic thermal data from existing real or synthetic visible face datasets. The generated pairs are used to adapt a pretrained visible-spectrum face recognition model into a CFR model. Unlike synthesis-based approaches that require image translation at test time, the proposed method shifts generation to the training stage and performs inference with a single forward pass through the adapted recognition model. Under the same MCXFace real-pair protocol, PACT improves over the evaluated CFR adaptation baselines, isolating the effect of the proposed adaptation objective. Training PACT on larger generated paired datasets provides additional improvements over both the unadapted model and the real-pair PACT configuration. Cross-database evaluation on the Tufts dataset provides evidence that the learned representation transfers to an unseen database. The source code and trained models will be made publicly available.
\end{abstract}

\section{Introduction}
Face recognition (FR) has evolved as a highly accurate and convenient modality for biometric authentication  \cite{kim202650}. While conventional visible-spectrum FR is easy to use, it often fails under poor illumination, at night, or in unconstrained environments. Thermal imaging offers an effective alternative since it captures heat emitted by the human face rather than reflected light. Modern thermal infrared sensors typically operate in the medium-wave infrared (MWIR) band of $3$--$5~\mu\mathrm{m}$ and the long-wave infrared (LWIR) band of $7$--$14~\mu\mathrm{m}$, enabling face imaging even under low-light or no-light conditions \cite{poster2021large}. Due to improvements in sensor technology and reductions in cost, thermal face recognition has become increasingly useful for law enforcement, surveillance, border security, and healthcare applications \cite{wang2020novel,osia2017facial}.

While thermal imaging offers several advantages, making it compatible with existing visible-spectrum galleries, such as national identity databases, is not trivial. Visible-to-thermal face recognition, a form of cross-spectral face recognition (CFR) \cite{anghelone2025beyond}, offers a practical alternative. CFR focuses on matching face images across visible and thermal modalities, such as comparing a visible gallery image with a thermal probe image captured at night. This eliminates the need for separate thermal enrollment, allowing visible-spectrum galleries from legacy systems to be reused and thereby augmenting existing recognition capabilities.

Despite its usefulness, visible-to-thermal face recognition remains challenging due to the large modality gap between visible and thermal images~\cite{poster2021large,di2021multi}. This challenge is further compounded by the scarcity of large-scale paired visible--thermal datasets, which are typically smaller and less diverse than visible-spectrum face datasets (as their collection requires specialized sensors and synchronization across modalities)~\cite{hu2016polarimetric,panetta2018comprehensive}. As a result, models are often trained under limited-data conditions, leading to poor performance. 

In this work, we investigate the use of synthetic visible--thermal face data to improve cross-spectral face recognition under limited real-data conditions. We propose a synthetic data generation pipeline and a cross-modal adaptation strategy, PACT (Preservation-Aware Cross-Spectral Tuning), that integrates paired visible--synthetic thermal samples into the learning process. By generating synthetic thermal counterparts for visible face images, the proposed approach provides additional supervision for learning cross-modal identity representations without requiring large-scale real paired visible--thermal data. We further show that this adaptation improves over existing CFR methods even under an identical real-data protocol and that the learned representation transfers to an unseen thermal database.

More broadly, we view synthetic cross-modal generation as a mechanism for \emph{paired-supervision amplification}: a limited set of real paired samples are used to learn the cross-modal mapping, which then converts large scale single-modality data into scalable paired supervision for representation learning. This formulation showcases a general strategy for cross-modal learning in domains where acquiring paired data is substantially more expensive than acquiring data in a single modality.

The main contributions of this work are listed below:
\begin{itemize}
\item We formulate visible–thermal generation as paired-data amplification: a diffusion model adapted on limited real pairs constructs larger paired adaptation sets from external visible-face images.
\item We introduce a cross-modal adaptation strategy, PACT (Preservation-Aware Cross-Spectral Tuning), that combines symmetric thermal--visible contrastive alignment with visible-domain preservation regularization.
\item We evaluate generated adaptation sets derived from both real and synthetic visible sources and assess transfer to the unseen Tufts database.
\item Synthesis is performed only during training, so deployed recognition requires a single model forward pass without any image translation, making deployment-friendly inference possible.
\end{itemize}
Finally, to support reproducibility and further extensions, we will publicly release the source code and trained models\footnote{\url{https://www.idiap.ch/paper/synthermface/}}.
\section{Related Work}
\begin{figure*}[t!]
  \centering
  \includegraphics[width=0.98\linewidth]{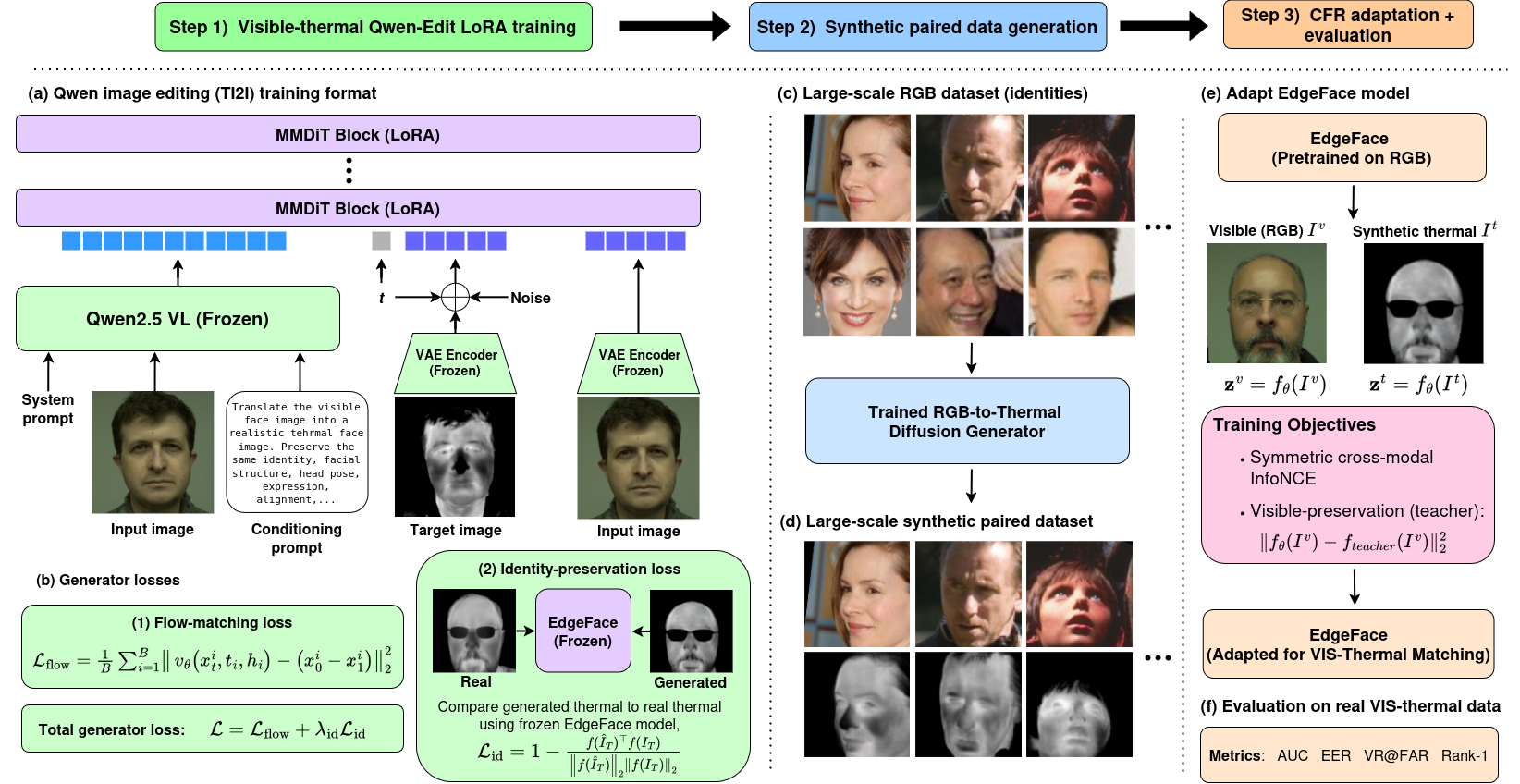}
  \caption{Overview of the proposed SynThermFace framework. A visible-to-thermal generation model is first fine-tuned using a limited set of paired visible-thermal images. The adapted generator is then used to synthesize thermal images for a larger set of identities, producing paired visible--synthetic thermal data for training the CFR model.}
  \label{fig:framework}
\end{figure*}

\textbf{Heterogeneous Face Recognition:} Heterogeneous face recognition (HFR), including cross-spectral face recognition (CFR), addresses face matching across different sensing modalities, such as VIS, NIR, and thermal imaging. Its main challenge is the modality gap, where images from different domains follow distinct distributions. Existing HFR methods can be broadly grouped into three categories. First, modality-invariant feature learning aims to extract representations that remain stable across domains, ranging from early handcrafted descriptors such as LBP and SIFT \cite{liao2009heterogeneous,klare2010matching} to deep CNN-based invariant representations \cite{he2017learning,he2018wasserstein}. Second, common-space projection methods reduce the modality gap by mapping features into a shared latent space using techniques such as CCA, PLS, coupled regression, or deep domain-invariant architectures \cite{yi2007face,sharma2011bypassing,george2024heterogeneous}. A third class of methods, called synthesis-based approaches, translates images from one modality to another, often into the visible domain, enabling the use of standard FR models. These methods evolved from patch-based reconstruction and manifold learning to GAN-based image translation frameworks such as CycleGAN \cite{wang2008face,zhuUnpairedImagetoImageTranslation2017}.
DiffTV \cite{lin2024difftv} proposes a latent diffusion model for identity-preserved thermal-to-visible face translation, using heterogeneous feature alignment and dual-stage conditioning to better preserve identity details, including facial structure and skin color. However, it is proposed as an image translation method for CFR.

\textbf{Synthetic Data for Face Recognition:} In conventional visible-spectrum face recognition, synthetic data is often used as an alternative or supplement to large-scale real face datasets. Recent works have explored synthetic face datasets to address the legal, privacy, and ethical challenges of collecting real facial data. Early works such as SynFace \cite{qiu2021synface} and SFace \cite{boutros2022sface} leveraged GAN-based models to generate identity-preserving synthetic faces. More recent approaches employ diffusion models, such as IDiff-Face \cite{boutros2023idiff}, or hybrid GAN--diffusion pipelines such as GANDiffFace \cite{melzi2023gandiffface}, to improve identity diversity and image realism. Other methods focus on disentangling and controlling identity representations in latent spaces, including ExFaceGAN \cite{boutros2023exfacegan}, IDNet \cite{kolf2023identity}, and Digi2Real \cite{george2025digi2real}. Additionally, synthetic data has been used for efficient model training and hard-sample generation \cite{shahreza2023synthdistill}, while advanced latent-space sampling strategies such as DisCo \cite{geissbuhler2024synthetic} further enhance identity diversity and intra-class variation. Although models trained solely on synthetic data still typically underperform those trained on real data, this performance gap has been steadily decreasing as generative models improve in realism, diversity, and identity consistency.
 
\textbf{Cross-Spectral Synthetic Data:} In \cite{farooq2026thermvision}, authors proposed ThermVision-DB, a synthetic long-wave infrared (LWIR) thermal face dataset generated using a FLUX-LoRA diffusion model and LivePortrait-based image-to-video retargeting. The dataset provides synthetic thermal facial images and videos with controlled identity, gender and other attributes, targeting privacy-preserving thermal face analysis. Although the work is relevant to cross-spectral synthetic data research, it does not demonstrate paired RGB--thermal model training or cross-spectral translation; instead, the generative pipeline is trained and evaluated within the thermal domain. The paper also does not discuss whether synthetic data can replace or improve over real data in downstream tasks. In \cite{tomavsevic2024generating}, authors introduced an identity-conditioned dual-branch StyleGAN2 framework for generating aligned visible (VIS) and near-infrared (NIR) synthetic face images for privacy-preserving face recognition. The method includes a Privacy and Diversity filter that removes samples matching real identities or previously generated identities while improving identity separability and intra-identity diversity. This work explicitly trains and evaluates recognition models using synthetic cross-spectral VIS--NIR data, showing that multispectral synthetic training can improve performance even on visible-spectrum benchmarks. However, they do not address the more challenging thermal images. T-FAKE \cite{flotho2025t} creates 200,000 synthetic thermal faces with sparse and dense landmarks by thermalizing synthetic RGB faces using paired reconstruction, Wasserstein patch-distribution matching, and region-specific temperature regularization. However, it targets facial landmarking rather than identity preservation or cross-spectral recognition.

\textbf{Motivation:} The main motivation for synthetic data in FR comes from privacy, consent, and regulatory concerns, whereas in cross-spectral face recognition the motivation is more pragmatic: \textit{paired visible--thermal data are scarce}, expensive, and difficult to scale.

Unlike RGB face images, which can be collected or web-scraped at large scale, cross-spectral datasets require specialized sensors and synchronized capture across modalities, resulting in limited identity and diversity. Synthetic data therefore offers a practical way to generate a large-scale paired visible--thermal dataset, expand diversity, and reduce dependence on costly real data collection. Unlike image-translation approaches for heterogeneous face recognition, which translate thermal or other non-RGB inputs into visible images at inference time \cite{mallat2019cross,zhang2017generative}, our approach shifts the generative cost to the training stage by generating a large-scale paired visible--thermal dataset from available visible face data and using it to fine-tune cross-spectral recognition models. This removes the need for a costly translation step for every test sample, reducing inference to a single forward pass through the recognition model and making the inference stage more practical.

\section{Proposed Method}

As discussed in the previous section, our main objective is to develop a high-performing visible-to-thermal face recognition model. To achieve this, we require paired visible--thermal data for training the cross-spectral face recognition (CFR) model. Our approach consists of three tightly coupled stages. First, we train a diffusion-based visible-to-thermal translation model using a small paired visible--thermal dataset. Second, we apply the trained translation model to large-scale real or synthetic visible face datasets, generating a synthetic thermal counterpart for each visible image and thereby constructing paired visible--synthetic thermal training data. Finally, we fine-tune a face recognition model on this synthetic paired data so that visible and thermal samples of the same identity are aligned in the feature space. The resulting model is then benchmarked for cross-spectral recognition against baseline models and models trained using real paired data, and its generalization is further assessed on an unseen thermal database. The overall framework is shown in Fig. \ref{fig:framework}. The details of each stage are elaborated in this section.

\subsection{Diffusion Model}

The first stage trains a visible-to-thermal generator using a limited set of real paired visible--thermal images. The trained generator is then applied to large-scale visible face datasets to synthesize thermal counterparts in the second stage, constructing paired visible--synthetic thermal data for CFR adaptation. 

Our objective in the first stage is to learn a visible-to-thermal translation model that can generate a thermal-like counterpart for an arbitrary visible face image while retaining the source image’s facial geometry, pose, and identity-relevant information. We formulate this as an image-editing problem and build on Qwen-Image-Edit \cite{wu2025qwenimagetechnicalreport}, a large-scale editing model based on a Multimodal Diffusion Transformer (MMDiT). The model couples a Qwen2.5-VL multimodal encoder that extracts high-level semantic conditioning with a variational autoencoder (VAE) that provides low-level appearance features, and is trained with a flow-matching objective in the VAE latent space. 

Let $x_0 = \mathcal{E}(I_T)$ denote the latent of the target thermal image obtained from the VAE encoder, and let $x_1  \sim \mathcal{N}(0, I)$ be a sampled noise vector. Following the rectified-flow formulation of Qwen-Image \cite{wu2025qwenimagetechnicalreport}, for each training sample a timestep $t \in [0, 1]$ is sampled from a logit-normal distribution, and the intermediate latent and its target velocity are defined as
\begin{equation}
    x_t = t \, x_0 + (1-t) \, x_1, \qquad v_t = x_0 - x_1 ,
\end{equation}

respectively. Given the user input $S$, which comprises a text prompt combined with an image, the multimodal encoder $\phi$ (Qwen2.5-VL) is used to obtain the guidance latent $h = \phi(S)$. Then, the transformer $v_\theta$ is trained to predict the target velocity, and over a mini-batch of size $B$, the flow-matching loss is given by  
\begin{equation}
    \mathcal{L}_{\mathrm{flow}} = \frac{1}{B} \sum_{i=1}^{B}
    \bigl\lVert\, v_\theta\!\left(x_t^{i}, t_i, h_i\right)
    - \left(x_0^{i} - x_1^{i}\right) \bigr\rVert_2^2 ,
\end{equation}

where, for each sample $i$ in the batch, $x_t^{i}$ is the noised latent obtained by interpolating between the target latent $x_0^{i}$ and the sampled noise $x_1^{i}$ at timestep $t_i$, and $h_i=\phi(S_i)$ is the conditioning embedding extracted by the multimodal encoder.

For editing, the source image is encoded through both pathways and jointly fed into the transformer as conditioning signals. This conditioning allows for the structure and layout of the original image to be preserved while appearance can be modified, which is well suited to our task of visible-to-thermal translation where the facial geometry and pose of the input must be retained and only the imaging modality should change. 

\textbf{Low Rank Adaptation:} Fully fine-tuning a 20B-parameter editing model on a small paired dataset is prone to overfitting, so we adapt it with Low-Rank Adaptation (LoRA) \cite{DBLP:journals/corr/abs-2106-09685}. We insert trainable low-rank updates into the attention projections and the MLP and modulation layers of both the image and text streams of the MMDiT, while keeping the VAE and Qwen2.5-VL encoder frozen. This restricts adaptation to a small fraction of the backbone, preserving the generative capabilities of the pretrained model while making fine-tuning feasible on our limited data. 

\textbf{Identity Preservation:} In addition to the standard losses used in Qwen-Edit, we introduce an identity-preservation loss that encourages generated images to retain the subject identity. Since our model performs RGB-to-thermal translation, enforcing identity preservation ideally requires a face recognition model that is robust across the visible-thermal domain gap. However, such a model is not available in our current setup; developing such a robust cross-spectral FR model is itself the goal of this work.

Instead, we use a proxy identity loss computed within the thermal modality. Specifically, during training, we reconstruct the predicted clean thermal image from the noisy latent and velocity prediction at low-noise timesteps, decode it through the VAE, and compare it with the corresponding ground-truth thermal image of the same subject. The comparison is performed in the feature space of a frozen lightweight EdgeFace-Base model \cite{george2024edgeface}, using a cosine-distance loss between the two identity embeddings. This loss is applied only when the noise level is sufficiently low, so that the reconstructed image is meaningful, and is added to the standard flow-matching loss as a weighted auxiliary objective. This design is motivated by the observation that thermal images of the same identity form well-separated clusters within the thermal modality \cite{george2022prepended,george2024edgeface}. Although EdgeFace was originally trained on visible images, we use it only as a weak identity regularizer within paired thermal supervision.

\begin{equation}
\mathcal{L}_{\mathrm{id}}
=
1 -
\frac{
f(\hat{I}_{T})^{\top} f(I_{T})
}{
\left\| f(\hat{I}_{T}) \right\|_{2}
\left\| f(I_{T}) \right\|_{2}
}.
\end{equation}

where $\hat{I}_{T}$ denotes the generated thermal image, $I_{T}$ is the corresponding ground-truth thermal image, and $f(\cdot)$ represents the feature embedding extracted by the frozen EdgeFace-Base network.

The final training loss combines both terms:
\begin{equation}
\mathcal{L}
=
\mathcal{L}_{\mathrm{flow}}
+
\lambda_{\mathrm{id}} \mathcal{L}_{\mathrm{id}}.
\end{equation}

where $\mathcal{L}_{\mathrm{flow}}$ denotes the standard flow-matching loss used by Qwen-Edit, and $\lambda_{\mathrm{id}}$ controls the contribution of the identity-preservation term. $\lambda_{\mathrm{id}}$ is set to $0.1$ in our experiments.

\subsection{PACT: Preservation-Aware Cross-Spectral Tuning}
\label{subsec:cross_modal_adaptation}

In the third stage, we fine-tune face recognition models using paired
visible--thermal data. This paired data can be either real or synthetically
generated. However, directly training with paired data may not be optimal for
maximizing cross-modal performance, since the model must reduce the
visible--thermal modality gap without destroying the identity-discriminative
structure learned from large-scale visible-spectrum pretraining. Instead of
training a cross-spectral face recognition model from scratch, we start from a
lightweight EdgeFace~\cite{george2024edgeface} backbone pretrained on a
large-scale RGB face dataset.

We propose \textbf{PACT} (Preservation-Aware Cross-Spectral Tuning), a
cross-modal adaptation strategy that aligns visible and thermal embeddings
while regularizing them with the pretrained visible-domain representation. PACT combines
a symmetric thermal--visible contrastive loss with a visible-domain
preservation loss computed against a frozen copy of the original pretrained
model.

PACT shares the use of cross-modal alignment and frozen-teacher
preservation with prior lightweight adaptation methods such as
xEdgeFace~\cite{george2025xedgeface,xedgefacetbiom}. Whereas pairwise objectives optimize each anchor primarily against one paired counterpart, PACT assigns positive probability mass to every opposite-modality sample of the same identity in the mini-batch and jointly optimizes thermal-to-visible and visible-to-thermal retrieval. The preservation term separately limits drift from the pretrained visible-domain representation.

Let $f_{\theta}(\cdot)$ denote the pretrained EdgeFace encoder (it can be any FR model). Given a paired
thermal and visible sample $(x_i^t,x_i^v,y_i)$, where $y_i$ denotes its subject identity,
we extract $\ell_2$-normalized feature embeddings as
\begin{equation}
  \mathbf{z}_i^t =
  \frac{f_{\theta}(x_i^t)}
       {\lVert f_{\theta}(x_i^t)\rVert_2},
  \qquad
  \mathbf{z}_i^v =
  \frac{f_{\theta}(x_i^v)}
       {\lVert f_{\theta}(x_i^v)\rVert_2}.
\end{equation}

To adapt the pretrained RGB representation to the cross-modal setting, we
optimize a combination of complementary objectives. First, a symmetric, identity-aware, multi-positive cross-modal InfoNCE
loss~\cite{oord2018representation, miech2020end} encourages the model to assign high
similarity to visible and thermal samples belonging to the same subject while
assigning lower similarity to samples from different subjects.

For a mini-batch of size $B$, we define the scaled similarity between a thermal
anchor and a visible candidate as
\begin{equation}
  s_{ij}
  =
  \frac{{\mathbf{z}_i^t}^{\top}\mathbf{z}_j^v}{\tau},
\end{equation}
where $\tau$ is a temperature hyperparameter.

For each anchor $i$, let
\begin{equation}
  \mathcal{P}(i)
  =
  \left\{j \in \{1,\ldots,B\} \mid y_j = y_i\right\}
\end{equation}
denote the set of visible or thermal samples in the mini-batch that share the
anchor's identity. The paired sample is always included because
$i\in\mathcal{P}(i)$. When multiple samples of the same subject occur in a
mini-batch, they are all treated as positives. 

The directional contrastive terms are then given by
\begin{equation}
\begin{aligned}
  \mathcal{L}_{t \rightarrow v}
  &= -\frac{1}{B}\sum_{i=1}^{B}
  \log \frac{\sum_{j\in\mathcal{P}(i)}\exp(s_{ij})}{\sum_{j=1}^{B}\exp(s_{ij})}, \\
  \mathcal{L}_{v \rightarrow t}
  &= -\frac{1}{B}\sum_{i=1}^{B}
  \log \frac{\sum_{j\in\mathcal{P}(i)}\exp(s_{ji})}{\sum_{j=1}^{B}\exp(s_{ji})}.
\end{aligned}
\end{equation}

The symmetric contrastive loss is defined as their average:
\begin{equation}
  \mathcal{L}_{\mathrm{NCE}}
  = \frac{1}{2}\left( \mathcal{L}_{t \rightarrow v} + \mathcal{L}_{v \rightarrow t} \right).
\end{equation}

The two terms correspond to thermal-to-visible and visible-to-thermal
matching, respectively.

This objective maximizes the total cross-modal similarity mass assigned
to same-identity candidates in the mini-batch while reducing the mass assigned
to candidates belonging to different identities.

To mitigate catastrophic forgetting, PACT retains a frozen copy of the
pretrained model $f_{\theta_0}$ and
encourages consistency between the adapted and original visible-image embeddings through
a visible-domain preservation loss:
\begin{equation}
  \mathcal{L}_{\mathrm{preserve}}
  =
  1-\frac{1}{B}\sum_{i=1}^{B}
  \cos\!\left(
    \mathbf{z}_i^v,
    \mathbf{z}_{i,\mathrm{teacher}}^v
  \right),
\end{equation}
where $\mathbf{z}_{i,\mathrm{teacher}}^v$ denotes the embedding produced by
the frozen teacher network. This regularization is designed to discourage catastrophic forgetting relative to the frozen teacher, helping retain the discriminative structure obtained during large-scale RGB pretraining while making it possible to adapt to the thermal domain. 
The overall PACT objective is the weighted sum:
\begin{equation}
  \mathcal{L}_{\mathrm{PACT}}
  =
  \lambda_{\mathrm{NCE}}\mathcal{L}_{\mathrm{NCE}}
  +
  \lambda_{\mathrm{preserve}}\mathcal{L}_{\mathrm{preserve}},
  \label{eq:cfr_objective}
\end{equation}
where $\lambda_{\mathrm{NCE}}$ and $\lambda_{\mathrm{preserve}}$ balance
cross-modal alignment against visible-domain preservation.

\subsection{Implementation Details}

\begin{wrapfigure}{r}{0.5\textwidth}
    \centering
    \vspace{-\intextsep}
    \includegraphics[width=0.45\textwidth]{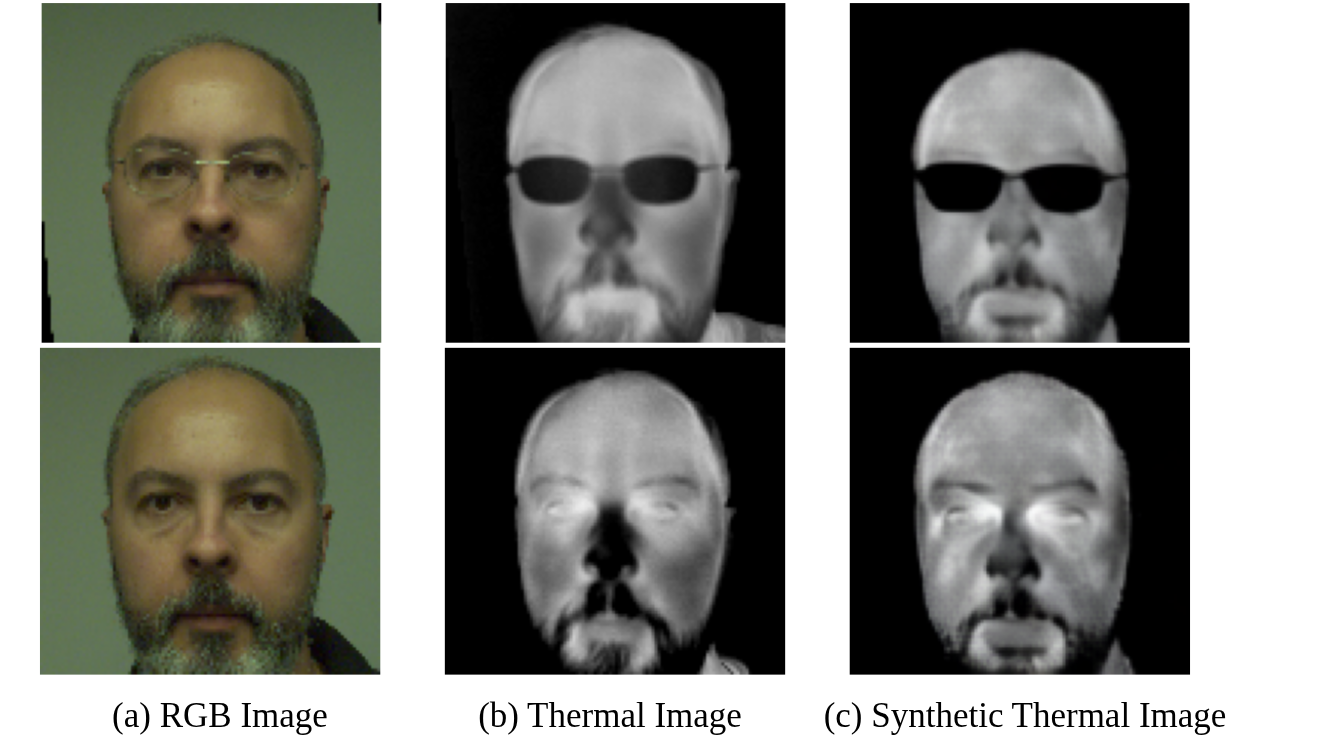}
    \caption{Examples of real visible and thermal images from MCXFace, along with the corresponding synthetic thermal images generated.}
    \label{fig:samples}
\end{wrapfigure}

The visible-to-thermal generator model is adapted from Qwen-Image-Edit-2511 using LoRA fine-tuning. Each training sample is a spatially aligned (visible, thermal) pair from the training set of the MCXFace VIS--THERMAL protocol. The visible image is provided as the editing condition and the thermal image as the generation target. Images are cropped and resized to $112 \times 112$. We fine-tune only the DiT component with LoRA rank 32, while using the pretrained Qwen2.5-VL text encoder and VAE. LoRA adapters are inserted into the attention projections, output projections, and selected image/text MLP and modulation layers. The model is trained for 20 epochs with a learning rate of $1 \times 10^{-4}$. Training is performed on an NVIDIA H100 GPU. During synthetic data generation, we use 20 denoising steps per image; generating one $112 \times 112$ thermal image takes approximately 10 seconds on the H100 GPU. Some sample images are shown in Fig. \ref{fig:samples}.

\begin{figure*}[htbp]
    \centering
    \includegraphics[
        width=0.98\linewidth,
        trim={0 0 0 1.2cm},
        clip
    ]{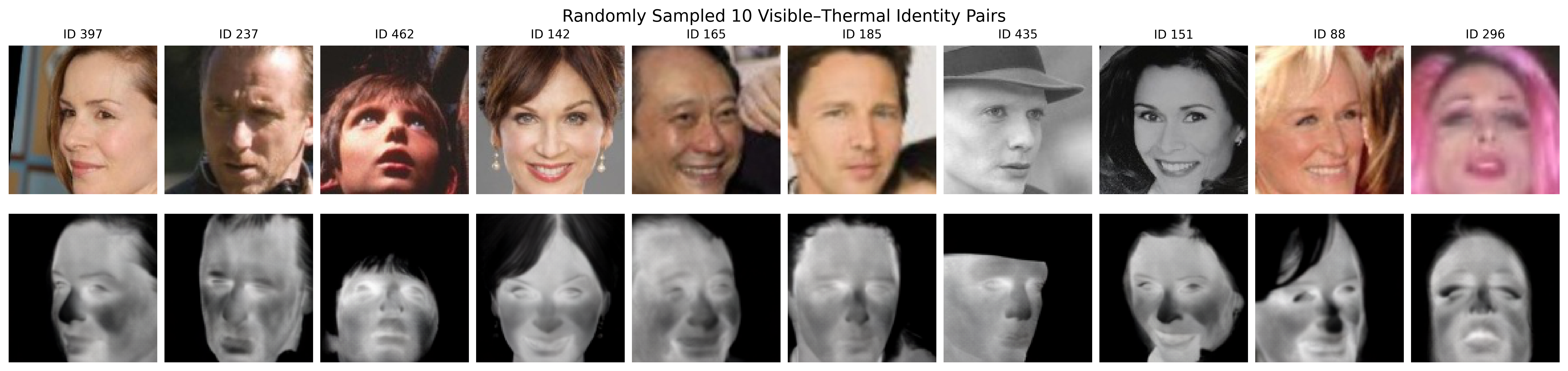}
    \caption{Samples showing real visible images from CASIA-WebFace and generated thermal samples.}
    \label{fig:samples_casia}
\end{figure*}

\begin{figure*}[htbp]
    \centering
    \includegraphics[
        width=0.98\linewidth,
        trim={0 0 0 1.2cm},
        clip
    ]{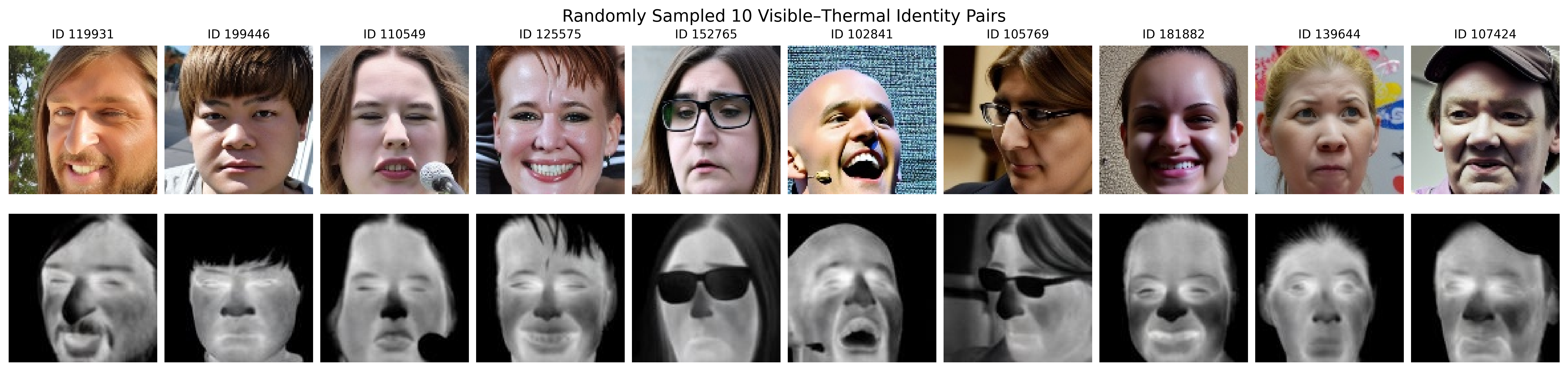}
    \caption{Samples showing synthetic visible images from Digi2Real and corresponding generated thermal samples.}
    \label{fig:samples_digi2real}
\end{figure*}

After training the visible-to-thermal translation model, we use CASIA-WebFace as the source of visible face images for large-scale synthetic data generation. Specifically, we sample up to 1000 identities with 20 images per identity and transform each visible image using the trained diffusion model. This produces paired visible--synthetic thermal datasets, referred to as CASIA-Synthetic-Thermal (Fig.~\ref{fig:samples_casia}) in the following sections, for cross-modal face recognition training. In addition, we use synthetic visible face images from Digi2Real as RGB inputs to generate corresponding thermal images, resulting in a fully synthetic paired dataset of 1000 identities referred to as Digi2Real-Synthetic-Thermal (Fig. \ref{fig:samples_digi2real}).

For the CFR adaptation stage, we fine-tune the pretrained EdgeFace-Base model using paired visible-synthetic thermal images generated by the proposed diffusion pipeline. Training is implemented in PyTorch using AdamW with a learning rate of $1 \times 10^{-5}$, batch size 128, and 20 epochs. Mini-batches are constructed by randomly shuffling paired samples. When multiple samples from the same identity occur in a batch, all corresponding cross-modal entries are treated as positives; otherwise, the objective reduces to the single-positive case. We optimize the weighted objective in Eq.~\ref{eq:cfr_objective}.  We fix $\lambda_{\mathrm{NCE}}=\lambda_{\mathrm{preserve}}=1$ a priori (equal weighting, as both losses act on normalized cosine similarities); Table~\ref{tab:nce_preserve_search} presents a post-hoc sensitivity analysis. The temperature $\tau=0.07$ and the identity-loss weight $\lambda_{\mathrm{id}}=0.1$ follow common practice and were fixed a priori. The $1{,}000$-identity setting follows the available generation budget; Table~\ref{tab:num_subjects_search} retrospectively examines sensitivity to overall training scale. All CFR adaptation experiments are trained on an NVIDIA RTX 3090 GPU. At inference time, only the adapted EdgeFace model is used, requiring a single forward pass without any image-translation module.

\section{Experiments}

This section describes the dataset generation process and the experimental setup used to evaluate the proposed approach.

\textbf{CFR Source Dataset:} We use the MCXFace dataset \cite{george2022prepended} for training and evaluating the model. MCXFace consists of images from 51 subjects captured across multiple channels, three acquisition sessions, and diverse illumination conditions. The dataset provides several modalities, including visible, thermal, and depth images. In this work, we focus on the visible-to-thermal setting, which represents one of the most challenging cross-modal scenarios in the dataset. The visible and thermal modalities are spatially registered, and the dataset includes predefined training and development sets with disjoint subject identities. For reproducibility, we conduct all experiments using the VIS-THERMAL protocol shipped with the dataset \cite{george2022prepended}.

\textbf{Metrics:} We evaluate performance using a set of standard metrics commonly adopted in prior literature. These include Area Under the Curve (AUC), Equal Error Rate (EER), Rank-1 identification rate, and Verification Rate measured at false acceptance rates of 0.1\%, 1\%, and 5\%.

\subsection{Baselines and Comparative Methods}

We compare the proposed method under four evaluation settings for VIS-THERMAL cross-spectral face matching:

\begin{itemize}
\item \textbf{Baseline Face Recognition Models:} We establish baseline performance by evaluating standard face recognition backbones on the VIS-THERMAL protocol using the \textit{dev} set.

\item \textbf{Comparative HFR Methods:} We evaluate representative heterogeneous face recognition (HFR) methods from the literature by training them on the training set and testing them on the \textit{dev} set.

\item \textbf{Synthetic Thermal Adaptation Data:} We assess the effectiveness of the proposed generation framework by training models using visible--synthetic thermal pairs generated from external visible face datasets and evaluating them on the \textit{dev} set.

\item \textbf{Isolating the Proposed Loss:} We evaluate EdgeFace adapted with the proposed PACT approach using only real MCXFace paired training data, to isolate the contribution of the adaptation loss from that of the synthetic data.

\end{itemize}

The real-pair experiments provide the controlled comparison of adaptation methods under the same MCXFace source-data protocol. The generated-data experiments evaluate the complete SynThermFace pipeline and should not be interpreted as data-matched comparisons against methods trained only on the smaller real MCXFace set.

\textbf{Baseline Methods:} To compare the effectiveness of the proposed approach, we first establish baseline results for VIS-THERMAL cross-spectral matching on the MCXFace dataset. We use the AdaFace model \cite{kim2022adaface}, which is based on the IResNet100 architecture and trained on WebFace12M \cite{zhu2022webface260m}, as well as EdgeFace \cite{george2024edgeface} (base model), which is a much more lightweight convolutional-transformer hybrid model, also trained on WebFace12M \cite{zhu2022webface260m}. These models cover both CNN and CNN-transformer hybrid architectures with high and low computational budgets. Note that both models are trained for visible-spectrum face recognition and are not adapted for cross-spectral face recognition.

\textbf{CFR Methods:} To ensure a fair comparison with existing cross-spectral face recognition methods, we include several CFR approaches from the literature and adapt them according to the MCXFace training protocol. Since evaluation is performed on a separate development set, this allows us to clearly assess the effect of model adaptation. Domain-Invariant Units (DIU) \cite{george2024heterogeneous} introduce a strategy for adapting pretrained face recognition models by fine-tuning only the lower layers to reduce modality dependence. This method is built on the AdaFace model. Prepended Domain Transformer (PDT) \cite{george2022prepended} adds a lightweight trainable prepended module for the target modality, making it easy to convert an existing face recognition model into a cross-spectral model. More recently, xEdgeFace \cite{george2025xedgeface,xedgefacetbiom} proposed an efficient adaptation strategy for lightweight face recognition models by tuning the layer normalization parameters and lower layers, achieving strong performance with low computational cost.

\textbf{Proposed Method:} In our proposed PACT approach, EdgeFace is adapted using the generated CASIA-Synthetic-Thermal or Digi2Real-Synthetic-Thermal dataset according to the cross-modal adaptation procedure described in Section~\ref{subsec:cross_modal_adaptation}.

\subsection{Experimental Results}

All methods are evaluated using the MCXFace VIS--THERMAL protocol. The training split is used for generator adaptation and, where applicable, CFR model training. Evaluation is performed only on the development split, whose identities are disjoint from the training split and are not used in training of any component for a fair evaluation.

Table~\ref{tab:model_comparison} compares baseline visible-spectrum face recognition models, existing CFR adaptation methods, and the proposed PACT adaptation scheme on the visible--thermal protocol. The baseline models perform poorly in cross-spectral matching, with EERs of 14.79\% and 23.06\% for AdaFace and EdgeFace, respectively, highlighting the large modality gap between visible and thermal face images. Existing CFR methods substantially improve performance through cross-modal adaptation. Among them, DIU achieves an EER of 3.73\%, while xEdgeFace obtains a similar EER of 3.76\% with a lightweight backbone.

PACT fine-tuned only on real paired MCXFace data already improves over these CFR baselines under an identical real-data protocol, which isolates the contribution of the proposed PACT loss from that of the synthetic data, achieving an EER of 3.04\% and a Rank-1 accuracy of 96.49\%. Notably, PACT fine-tuned on real data also achieves strong low-FAR verification performance.

When PACT is fine-tuned on fully synthetic visible--thermal pairs (data from Digi2Real-Synthetic-Thermal), performance improves further across all metrics. The fully synthetic setting reduces the EER to 1.20\%, improves Rank-1 accuracy to 99.50\%, and achieves 99.75\% VR@5\%, 98.75\% VR@1\%, and 92.48\% VR@0.1\%. The partially synthetic CASIA-Synthetic-Thermal setting performs best overall, with an EER of 0.99\% and Rank-1 accuracy of 99.75\%. These results show that the generated pairs provide useful supervision within the evaluated PACT pipeline. Importantly, the improvement over the real-pair PACT configuration shows that the larger generated adaptation set provides additional useful supervision. 

Although the generated PACT adaptation set is larger, the generator’s cross-modal supervision originates from the limited MCXFace training split. The proposed pipeline therefore reuses limited real supervision to construct a larger adaptation dataset.

The two synthetic settings differ only in their visible source: CASIA-Synthetic-Thermal pairs \emph{real} CASIA-WebFace visible images with generated thermal images (partially synthetic), whereas Digi2Real-Synthetic-Thermal pairs \emph{synthetic} Digi2Real visible images with generated thermal images (fully synthetic). As shown in Table~\ref{tab:model_comparison}, the two visible-source settings perform similarly, with a small numerical advantage for CASIA-Synthetic-Thermal on MCXFace. We clarify that ``fully synthetic'' here refers to the paired adaptation set: the generator itself is still fine-tuned on real paired MCXFace data, so the pipeline still relies on some real supervision.

\begin{table*}[t]
\centering
\caption{Performance comparison of different FR and CFR models. Higher is better for all metrics except EER.}
\label{tab:model_comparison}
\resizebox{0.99\linewidth}{!}{%
\begin{tabular}{lcccccc}
\toprule
Model
& AUC (\%) $\uparrow$
& EER (\%) $\downarrow$
& VR@5\% (\%) $\uparrow$
& VR@1\% (\%) $\uparrow$
& VR@0.1\% (\%) $\uparrow$
& R1 (\%) $\uparrow$ \\
\midrule
AdaFace~\cite{kim2022adaface}
& 93.89 & 14.79 & 76.19 & 52.88 & 38.35 & 65.16 \\

EdgeFace~\cite{george2024edgeface}
& 86.08 & 23.06 & 48.37 & 28.82 & 13.03 & 40.10 \\

\midrule
DIU~\cite{george2024heterogeneous}
& 99.22 & 3.73 & 96.99 & 80.70 & 41.60 & 90.73 \\

xEdgeFace~\cite{george2025xedgeface}
& 99.34 & 3.76 & 97.49 & 86.22 & 64.41 & 90.48 \\

PDT~\cite{george2022prepended}
& 99.00 & 6.00 & 93.23 & 86.22 & 72.18 & 89.47 \\

\midrule
PACT (MCXFace-Real)
& 99.65 & 3.04 & 98.75 & 90.73 & 72.68 & 96.49 \\
\midrule

PACT (Digi2Real-Synthetic-Thermal)
& 99.92 & 1.20 & \textbf{99.75} & 98.75 & 92.48 & 99.50 \\

PACT (CASIA-Synthetic-Thermal)
& \textbf{99.94} & \textbf{0.99} & \textbf{99.75}
& \textbf{99.25} & \textbf{93.48} & \textbf{99.75} \\
\bottomrule
\end{tabular}
}
\end{table*}

\subsection{Ablations}

In this section, we present ablation studies to analyze the contribution of different components of the proposed pipeline and to better understand the factors influencing cross-spectral recognition performance.

\textbf{Sensitivity to Loss Weights:}
The weights of the NCE and preservation losses control the trade-off between cross-modal alignment and retention of the pretrained visible-domain embedding space. To study their effect, we vary the corresponding loss weights and report the results in Table \ref{tab:nce_preserve_search}. When both weights are set to zero, the model reduces to the unadapted EdgeFace baseline, resulting in poor visible--thermal matching performance. Introducing the NCE loss leads to a substantial improvement, confirming the importance of contrastive cross-modal alignment. Adding the preservation loss further improves the reported metrics, consistent with its intended role of limiting drift from the pretrained identity representation.  Performance is stable around balanced non-zero weights (EER $\approx 1.0\%$ for $\lambda_{\mathrm{NCE}},\lambda_{\mathrm{preserve}}\in\{0.5,1.0\}$), with some metrics favoring $1.0/0.5$; we adopt equal $1.0/1.0$ weighting.

\begin{table*}[t]
\centering
\renewcommand{\arraystretch}{1.2}
\setlength{\tabcolsep}{3pt}
\begin{minipage}[t]{0.49\linewidth}
\centering
\caption{Sensitivity to the NCE and preservation loss weights.}
\label{tab:nce_preserve_search}
\scriptsize
\begin{tabular}{cccccc}
\toprule
$\lambda_{\mathrm{NCE}}$ & $\lambda_{\mathrm{Preserve}}$ & AUC $\uparrow$ & EER $\downarrow$ & VR@1\% $\uparrow$ & VR@0.1\% $\uparrow$ \\
\midrule
0.00 & 0.00 & 86.08 & 23.06 & 28.82 & 13.03 \\
0.00 & 0.50 & 86.30 & 23.06 & 30.08 & 12.78 \\
0.00 & 1.00 & 86.31 & 22.81 & 29.82 & 12.78 \\
\midrule
0.50 & 0.00 & 99.79 & 2.46 & 93.48 & 74.19 \\
0.50 & 0.50 & 99.95 & 1.00 & 99.00 & 93.23 \\
0.50 & 1.00 & 99.91 & 1.20 & 98.50 & 92.23 \\
\midrule
1.00 & 0.00 & 99.83 & 2.01 & 95.74 & 76.19 \\
1.00 & 0.50 & \textbf{99.96} & 1.00 & 99.00 & \textbf{94.24} \\
1.00 & 1.00 & 99.94 & \textbf{0.99} & \textbf{99.25} & 93.48 \\
\bottomrule
\end{tabular}
\end{minipage}\hfill
\begin{minipage}[t]{0.49\linewidth}
\centering
\caption{Effect of the number of synthetic training identities.}
\label{tab:num_subjects_search}
\scriptsize
\begin{tabular}{ccccc}
\toprule
IDs & AUC $\uparrow$ & EER $\downarrow$ & VR@1\% $\uparrow$ & VR@0.1\% $\uparrow$ \\
\midrule
10   & 94.03 & 13.50 & 49.37 & 26.57 \\
50   & 99.44 &  3.51 & 87.97 & 67.17 \\
100  & 99.76 &  2.46 & 93.98 & 71.93 \\
200  & 99.82 &  1.98 & 95.49 & 80.20 \\
500  & 99.91 &  1.29 & 98.25 & 87.97 \\
1000 & \textbf{99.94} & \textbf{0.99} & \textbf{99.25} & \textbf{93.48} \\
\bottomrule
\end{tabular}
\end{minipage}
\end{table*}

\textbf{Effect of the Number of Identities:}
We further analyze the effect of the number of synthetic training identities used for cross-modal adaptation. For this experiment, we generate visible-synthetic thermal pairs from CASIA-WebFace using different numbers of identities, while keeping the NCE and preservation loss weights fixed at 1.0. The results are reported in Table \ref{tab:num_subjects_search}. Performance improves as the overall generated adaptation set is scaled from 10 to 1,000 identities. The marginal gain becomes smaller beyond 500 identities, although identity count, image count, and optimization budget vary jointly.

\subsection{Cross-Database Evaluation on Tufts Face Dataset}
\label{subsec:cross_database_tufts}

To assess whether the proposed adaptation overfits to MCXFace-specific data statistics, we evaluate the trained models on the Tufts Face Dataset~\cite{panetta2018comprehensive}. Tufts contains multi-modal face images from 113 identities (39 males and 74 females). We follow the VIS--Thermal protocol in~\cite{fu2021dvg} and use Tufts only for cross-database testing.

Table~\ref{tab:tufts_cross_database} compares the untuned EdgeFace baseline, xEdgeFace fine-tuned on MCXFace, PACT fine-tuned on real MCXFace pairs, and PACT fine-tuned on synthetic (Digi2Real and CASIA) data. All adapted models improve over the baseline, but PACT gives stronger transfer than xEdgeFace. PACT fine-tuned on real MCXFace pairs reduces EER from 43.41\% to 18.21\%, while PACT fine-tuned on fully synthetic data further reduces EER to 13.91\% and improves Rank-1 accuracy to 56.37\%.

These results provide evidence that the learned adaptation transfers beyond MCXFace-specific identities and acquisition conditions. However, the substantial performance gap between MCXFace and Tufts shows that sensor and database shift remains an open challenge. The Digi2Real-derived setting achieves the lowest EER and highest AUC and verification rates, while the CASIA-derived setting achieves the highest Rank-1 accuracy.

\begin{table}[t]
\centering
\caption{Cross-database VIS-Thermal evaluation on the Tufts Face Dataset. Higher is better for all metrics except EER.}
\label{tab:tufts_cross_database}
\resizebox{\linewidth}{!}{%
\begin{tabular}{lccccc}
\toprule
Model & AUC (\%) $\uparrow$ & EER (\%) $\downarrow$ & VR@5\% (\%) $\uparrow$ & VR@1\% (\%) $\uparrow$ & R1 (\%) $\uparrow$ \\
\midrule
EdgeFace~\cite{george2024edgeface} & 59.40 & 43.41 & 13.36 & 2.97 & 12.03 \\ \midrule
xEdgeFace~\cite{george2025xedgeface} (MCXFace) & 83.32 & 23.90 & 46.94 & 30.98 & 31.06 \\ \midrule
PACT (MCXFace-Real) & 90.27 & 18.21 & 66.23 & 49.72 & 48.83 \\ \midrule
PACT (Digi2Real-Synthetic-Thermal) & \textbf{93.49} & \textbf{13.91} & \textbf{75.70} & \textbf{58.81} & 56.37 \\
PACT (CASIA-Synthetic-Thermal) & 93.01 & 14.70 & 75.32 & 58.26 & \textbf{57.45} \\
\bottomrule
\end{tabular}
}
\end{table}

\subsection{Discussion}

The experiments support three conclusions. First, under the same real-pair protocol, PACT improves over the evaluated CFR adaptation methods. Second, training PACT on larger generated paired datasets yields additional gains over the limited real-pair configuration. Third, the improvements observed on Tufts provide evidence of cross-database transfer. The current experiments do not independently establish physical thermal realism or exact identity preservation; instead, the generated data are evaluated through their end-to-end utility for downstream cross-spectral face recognition.

Since large-scale real visible--thermal paired data are scarce and costly to acquire, the purpose of the proposed synthetic generation stage is to amplify limited real paired supervision into a substantially larger adaptation set. The results demonstrate the practical value of synthetic data for scaling cross-modal supervision beyond the available real pairs.

Taken together, the results indicate that generated visible–synthetic thermal pairs provide useful supervision for downstream cross-spectral adaptation. The thermal generator still depends on limited real paired MCXFace data, so the approach reduces rather than eliminates the need for real cross-modal supervision. The Digi2Real result suggests that synthetic visible images can serve as effective conditioning sources. However, because adaptation-set size, identity count, and training exposure vary jointly, the current experiments do not isolate the precise source of the improvement. The Tufts evaluation further shows that substantial sensor and database shift remains.

\subsection{Limitations}
While the proposed approach substantially improves performance, it has some limitations. The generated images may inherit biases and artifacts from the base diffusion model, the paired data used for fine-tuning, and the recognition model used for identity supervision. We further note that directly and quantitatively measuring the identity preservation of the generated thermal images is not trivial: verifying that a generated thermal face retains the identity of its visible source would itself require a reliable visible--thermal cross-spectral recognition model, which does not exist a priori and is precisely the goal of this work. A visible-trained recognizer is unreliable across the modality gap, while a thermal recognizer requires ground-truth thermal images that are available only for the small paired set and not for the large-scale generated data. We therefore assess the utility indirectly, through downstream cross-spectral recognition performance. In the current work, we evaluate the proposed approach only with EdgeFace, which serves as both the auxiliary identity supervisor during generator adaptation and the downstream recognition backbone, due to its lightweight design. Evaluating the approach with independent identity supervisors and additional, higher-capacity recognition architectures is left for future work. Future work will also scale the generated dataset to include more identities and richer pose, expression, demographic, and environmental variations, and extend evaluation to additional spectral bands and larger cross-database benchmarks.

\textbf{Ethical considerations:} The visible source images (CASIA-WebFace) are web-collected and are used here strictly for research; the released models and code are intended for research use only. Cross-spectral face recognition raises privacy and surveillance concerns, and generative pipelines may propagate demographic biases present in the source data. We encourage responsible use, adherence to dataset licenses and applicable regulations, and further study of demographic fairness prior to any deployment.

\section{Conclusions}

In this work, we introduce SynThermFace, a framework for amplifying limited real visible–thermal supervision through offline synthetic paired generation. PACT combines symmetric identity-aware cross-modal alignment with regularization toward a frozen visible-spectrum representation. Under the same MCXFace real-pair protocol, PACT improves over the evaluated CFR adaptation baselines. Training on larger generated adaptation sets provides further gains, while evaluation on Tufts provides evidence of transfer to an unseen database. These experiments establish the downstream utility of generated visible–synthetic thermal pairs. The source code and trained models will be made publicly available for enabling future extensions of the work.

\section*{Acknowledgment}
The project leading to this work has received funding from Frontex under the Frontex Research Grants Programme. Call for Proposals 2024/CFP/INNOVATE/01
Grant Agreement No. 2025/280. This work reflects only
the authors’ view. Neither the European Union nor Frontex
are responsible for any use that may be made of the information it contains. This research was also partly funded by the European Union project CarMen (Grant Agreement No.
101168325).
\bibliography{egbib}
\end{document}